# Markov Blanket Ranking using Kernel-based Conditional Dependence Measures

**Eric V. Strobl**                                                                 EVS17@PITT.EDU
**Shyam Visweswaran**                                                                SHV3@PITT.EDU
*Department of Biomedical Informatics*
*University of Pittsburgh School of Medicine*
*5607 Baum Boulevard*
*Pittsburgh, PA 15206, USA*

### Abstract

Developing feature selection algorithms that move beyond a pure correlational to a more causal analysis of observational data is an important problem in the sciences. Several algorithms attempt to do so by discovering the Markov blanket of a target, but they all contain a forward selection step which variables must pass in order to be included in the conditioning set. As a result, these algorithms may not consider all possible conditional multivariate combinations. We improve on this limitation by proposing a backward elimination method that uses a kernel-based conditional dependence measure to identify the Markov blanket in a fully multivariate fashion. The algorithm is easy to implement and compares favorably to other methods on synthetic and real datasets.

## 1 Introduction

Causality refers to a relation between a variable and another variable such that the latter variable is understood to be a consequence of the former. Three groups of methods have been described in the literature to infer causality from observational data. The most popular group includes conditional independence test methods such as PC (Spirtes et al., 2000) and FCI (Spirtes, 2001) that attempt to construct a graph representing all causal relationships in a dataset. The second group takes a more local approach by identifying the Markov blanket, or those variables that are conditionally independent on a target given the remaining variables; examples include IAMB (Tsamardinos & Aliferis, 2003), HITON-MB (Aliferis et al., 2003), and MMMB (Tsamardinos et al., 2006). The final group identifies pair-wise causal relationships by comparing the complexities of a forward and backward model such as LiNGAM (Shimizu et al., 2006) and additive noise models (Hoyer et al., 2008). However, to remain tractable, all of these methods do not consider all possible multivariate combinations. As a result, they may fail to identify subtle dependencies between variables.

A number of kernel-based methods have recently been developed that perform multivariate conditional dependence measurements in reproducing kernel Hilbert space (RKHS; Fukumizu et al., 2009; Zhang et al., 2011). In this paper, we take advantage of these methods by incorporating either one of two kernel-based conditional dependence measures (K-CDMs; Fukumizu et al., 2009; Zhang et al., 2011) in a backward elimination algorithm to identify the Markov blanket in a fully multivariate fashion. The rest of this paper is structured as follows. We first provide background on Bayesian networks in Section 2 and then discuss related work in Section 3. In Section 4, we describe the new algorithm that identifies the Markov blanket of a target by iteratively eliminating variables that minimize K-CDM. We finally provide results comparing the proposed algorithm with other feature ranking and subset selection methods in Section 5. Section 6 provides a brief conclusion.

## 2  Background

We denote random variables in upper case italics and sets of random variables in upper case bold italics. A Bayesian network is a probabilistic model that combines a directed acyclic graph (DAG) with parameters to represent a joint probability distribution over a set of random variables. Specifically, a DAG contains a node for every variable in the dataset, and an edge between a pair of nodes $R$-$S$ is absent if $R$ is independent of $S$ given $T$ for some $T$, and an edge $R$-$S$ is present if $R$ is dependent on $S$ given $T$ for all $T$ (Friedman & Koller, 2009). The absence of edges in a DAG can be determined by performing tests of conditional independence. Two variables $R$ and $S$ are conditionally independent given a third variable $T$ if and only if the value of $R$ provides no information about the value of $S$ and vice versa given the value of $T$. In mathematical notation, $R \perp\!\!\!\perp S|T$.

Let $Y$ denote the target variable and $\boldsymbol{X}$ denote all other variables excluding $Y$ in a dataset. The Markov blanket of $Y$, denoted by $\boldsymbol{MB}(Y)$, is a subset of $\boldsymbol{X}$ that includes $Y$'s parent, child and spousal nodes. $\boldsymbol{MB}(Y)$ can be identified by showing that a target node is conditionally independent of all other nodes given its parents, children and spouses:

$$Y \perp\!\!\!\perp \{\boldsymbol{X}\backslash\boldsymbol{MB}(Y)\}|\boldsymbol{MB}(Y) \Leftrightarrow Y \perp\!\!\!\perp \boldsymbol{X}|\boldsymbol{MB}(Y). \tag{1}$$

In this paper, we assess conditional dependence between arbitrary distributions within RKHSs. Specifically, we map $\boldsymbol{X}$ and $Y$ into RKHSs $\mathcal{F}$ and $\mathcal{G}$ respectively using two positive semidefinite kernels $K_{\boldsymbol{X}}: \mathcal{X} \times \mathcal{X} \to \mathbb{R}$ and $K_Y: \mathcal{Y} \times \mathcal{Y} \to \mathbb{R}$. There then exists a conditional cross-covariance operator $\Sigma_{YY|\boldsymbol{X}}: \mathcal{G} \to \mathcal{G}$ for any function $g \in \mathcal{G}$ such that:

$$\langle g, \Sigma_{YY|\boldsymbol{X}} g\rangle_{\mathcal{G}} = \mathbb{E}_{\boldsymbol{X}}\left[Var_{Y|\boldsymbol{X}}[g(Y)|\boldsymbol{X}]\right], \tag{2}$$

which represents the residual errors of predicting $g(Y)$ with $\boldsymbol{X}$ (Fukumizu et al., 2009).

Let $\boldsymbol{X}_S$ be a subset of $\boldsymbol{X}$. Then, the conditional cross-covariance operator exhibits the following property: $\Sigma_{YY|\boldsymbol{X}_S} \geq \Sigma_{YY|\boldsymbol{X}}$, where the order is determined by the trace operator, and the equality holds when $\boldsymbol{X}_S$ includes $\boldsymbol{MB}(Y)$ so that $Y \perp\!\!\!\perp \boldsymbol{X}|\boldsymbol{X}_S$.

Empirically, we can compute the kernel matrices $K_{\boldsymbol{X}_S}$ and $K_Y$ from a sample size of $n$ drawn i.i.d. from the distribution $P(\boldsymbol{X}, Y)$. The trace of the empirical conditional cross-covariance operator is then defined by:

$$M^1 = tr\big(G_Y(G_{\boldsymbol{X}_S} + n\varepsilon I_n)^{-1}\big), \tag{3}$$

where $G_{\boldsymbol{X}_S} = \left(I_n - \frac{1}{n}\mathbf{1}_n\mathbf{1}_n^T\right) K_{\boldsymbol{X}_S} \left(I_n - \frac{1}{n}\mathbf{1}_n\mathbf{1}_n^T\right)$ with $n$ representing sample size, $I_n$ an $n \times n$ identity matrix, and $\mathbf{1}_n$ a vector of ones. The regularization term $\varepsilon \to 0$ is added for the inversion. A similar measure proposed by Zhang et al. (2011; Equation 12) is based on eigenvalue decompositions of centralized kernel matrices:

$$M^2 = tr\big(T_{\boldsymbol{X}_S} G_Y T_{\boldsymbol{X}_S}\big), \tag{4}$$

where $T_{\boldsymbol{X}_S} = \varepsilon\big(G_{\boldsymbol{X}_S} + \varepsilon I_n\big)^{-1}$. Unlike $M^1$, this new measure was developed so that the authors could create a test of conditional independence using a statistic shown in their Equation 13 whose null distribution is approximated by a gamma distribution. Note that both $Y$ and $\boldsymbol{X}_S$ can each be multivariate with either of the two K-CDMs. Moreover, both K-CDMs do not make assumptions about the data distributions of $Y$ and $\boldsymbol{X}_S$ or their functional relationship.

## 3   Related Work

The Hilbert Schmidt Independence Criterion (HSIC; Gretton et al., 2005) is a sensitive measure of dependence between two kernels, where larger values denote a greater degree of dependence. Song et al. (2007) developed an algorithm called BAHSIC that uses HSIC for feature selection by embedding the target in the first kernel and the remaining variables in the second kernel; the algorithm then uses backward elimination to remove variables from the second kernel that maximize HSIC. In practice, the algorithm can detect subtle dependencies and help increase classification accuracy to a greater extent than many other feature selection algorithms.

HSIC unfortunately can have difficulty in detecting all of the variables in $MB(Y)$, since some of these variables may only show a weak association with the target. Measures of conditional dependence may instead be more useful in this regard. Nevertheless, correctly identifying the subset of variables to condition on can be difficult as the number of possible subsets grows exponentially with the number of variables (Statnikov et al., 2013). Markov blanket discovery algorithms including IAMB, HITON-MB, and MMMB thus incorporate a forward selection phase, where variables are required to display an association to the target before being included in the conditioning set. For example, the HITON-MB algorithm relies on a univariate association between the tested variable $R$ and the target $Y$. On the other hand, the IAMB and MMMB algorithms test the association between $R$ and $Y$ relative to a growing conditioning set of previously selected variables. In other words, both IAMB and MMMB initially rely on a univariate relationship with $Y$ but gradually become more multivariate. These forward selection strategies can be suboptimal because some variables may reveal a relationship with the target only when all the other variables in $MB(Y)$ are included in the conditioning set.

Several other limitations have been described in the literature. First, HITON-MB and MMMB may identify incorrect variables in the second step, since there are certain conditions under which variables *not* in $MB(Y)$ can enter $MB(Y)$ as described in Peña et al. (2006). Moreover, both these algorithms rely on HITON-PC and MMPC which also have shortcomings. The PC algorithms assume that if $A$ is not a member of the set of variables which are parents and children of $Y$, denoted by $pc(Y)$, then $Y \perp\!\!\!\perp A|B$ for some $B \subseteq pc(Y)$, so any node *not* in $pc(Y)$ is removed which is not always true. Second, Lou and Obradovic (2010) highlight that conditional independence testing may become unreliable with small sample sizes. As a result, they have instead promoted algorithms that rely on sensitive dependence measurements such as HSIC as opposed to tests in order to discover $MB(Y)$. However, in this paper, we will show that a new algorithm using Equation 3 or 4 can in fact perform very well by similarly avoiding statistical testing.

The main ideas used in this paper are motivated by the work of Fukumizu et al. (2009), in which the authors introduced a method of kernel dimension reduction using Equation 3. However, their method cannot be directly used to find $MB(Y)$, since it finds orthogonal projections of $X$ with respect to *kernel-induced feature* space. In this paper, we select variables with respect to *input* space to make the kernel-based conditional dimensionality reduction method more applicable to $MB(Y)$ discovery.

## 4   The Algorithm

### 4.1   The Main Idea

We discover $MB(Y)$ using backward elimination. First, consider measuring the conditional dependence of $Y$ and $X$ given $X_S$, where $X_S \leftarrow X$. Clearly, the conditional dependence measure is zero, since $X$ cannot explain $Y$ given itself. Next, consider removing a variable from the conditioning set $X_S$. Since a target is completely shielded from the other variables given its $MB(Y)$ by the definition of a Markov blanket, eliminating a variable in $MB(Y)$ from $X_S$ will cause the K-CDM to return a *larger* value (assuming enough sample size), since now $X$ can better explain $Y$ when $X_S$ is missing a variable in $MB(Y)$. In contrast, removing a variable *not* in $MB(Y)$ will make *no difference*, since the conditional dependence measure is still zero if $X_S$ contains $MB(Y)$. This process of successively testing the removal of a variable in the conditioning set $X_S$ and then permanently removing the variable that minimizes K-CDM is repeated until $X_S$ is empty.

## 4.2 Implementation

The proposed method is a feature ranking algorithm that performs backward elimination using a K-CDM. The pseudo-code for the method is shown in Algorithm 1, such that K-CDM is written as:

$$M^*(Y, X_S, \sigma),$$

which denotes $M^1$ or $M^2$ evaluated with $Y$, $X_S$, and $\sigma$ such that $\sigma$ is the set of kernel hyperparameters (if any).

---
**Algorithm 1: Backward Elimination**
1. **Input:** Target feature $Y$, non-target features $X$
2. **Output:** Non-target features in ascending order $X^\dagger$
3. $X_S \leftarrow X$
4. $X^\dagger \leftarrow \emptyset$
5. **repeat**
6. $\quad x \leftarrow \min_{X \in X_S} M^*(Y, \{X_S \setminus X\}, \sigma), \sigma \in \Xi$
7. $\quad X_S \leftarrow X_S \setminus X$
8. $\quad X^\dagger \leftarrow X^\dagger \cup X$
9. **until** $X_S = \emptyset$

---

The algorithm works as follows. It first computes K-CDM for every variable eliminated from the conditioning set $X_S$ using appropriate kernel hyperparameters $\sigma$ (if any) chosen with a user defined method $\Xi$. For example, the Gaussian sigma hyperparameter can be defined as the median distance between data points. The identified variable $X$ which minimizes K-CDM when removed is then permanently removed from $X_S$ and placed into $X^\dagger$. The above procedure is repeated until $X_S$ is empty. The underlying principle behind the algorithm is thus to find the variable combination that can best explain the dependence between $Y$ and $X$ by iteratively eliminating those variables that can least explain the dependence.

Note that the above procedure has some advantages over previous methods from the nature of directly performing backward elimination rather than first performing a forward selection step. First, the method considers *all* possible multivariate relationships in $MB(Y)$, since all variables in $MB(Y)$ are eliminated from $X_S$ *after* the other variables assuming sufficient sample size to detect the relationships. Second, the proposed algorithm outputs a ranking of variables defined by the relative amounts of conditional dependence across the entire dataset. As a result, the ranking represents the relative importance of each of the variables in $MB(Y)$.

---
**Algorithm 2: Forward Selection**
1. **Input:** Target feature $Y$, non-target features $X$
2. **Output:** Non-target features in descending order $X^\dagger$
3. $X_S \leftarrow X$
4. $X^\dagger \leftarrow \emptyset$
5. **repeat**
6. $\quad x \leftarrow \min_{X \in X_S} M^*(Y, \{X^\dagger \cup X\}, \sigma), \sigma \in \Xi$
7. $\quad X_S \leftarrow X_S \setminus X$
8. $\quad X^\dagger \leftarrow X^\dagger \cup X$
9. **until** $X_S = \emptyset$
---

The forward selection procedure (Algorithm 2) is faster and can be implemented by including variables in $X^\dagger$ in line 6 rather than removing variables from $X_S$. However, this method underperforms backward elimination in practice and is *not* guaranteed to return $MB(Y)$ in the infinite sample limit, since conditional dependence is not assessed within the context of the other variables in $X$. Also note that the output is in descending order in $X^\dagger$ instead of in ascending order.

### 4.3 Proof of Correctness

**Theorem.** The final variables in $X^\dagger$ from Algorithm 1 will include $MB(Y)$ under the assumptions that (1) K-CDM is defined by Equation 3 or 4, and (2) the dataset $\{X \cup Y\}$ has an infinite sample size and is drawn i.i.d. from a joint probability distribution faithful to a DAG.

**Proof.** First, a lower value returned from Equation 3 or 4 denotes a higher degree of conditional independence between $Y$ and $X$ given $X_S$ than a higher value by design. Second, $Y$ is conditionally independent of $X$ given $MB(Y)$ by the definition of a Markov blanket. As a result, K-CDM is guaranteed to return a higher value every time a variable in $MB(Y)$ is tested for removal in line 6 compared to a variable *not* in $MB(Y)$ assuming an infinite sample size, where the data points are drawn i.i.d. from a probability distribution faithful to a DAG. Then, if $X_S$ contains variables in and not in $MB(Y)$, a variable *not* in $MB(Y)$ will be eliminated *earlier* from $X_S$ in line 7. The variable eliminated from $X_S$ will then be placed into $X^\dagger$ in line 8. As a result, the final variables in $X^\dagger$ will include $MB(Y)$.□

### 4.4 Time Complexity

We assume that we remove $1 - \beta$ of $X_S$ at every iteration. Then, the $i^{th}$ iteration of Algorithm 1 takes $O(\beta^{i-1}dn^3)$ where $d$ represents the total number of variables and $n^3$ represents the inversion of the kernel when calculating K-CDM. Similarly, the $i^{th}$ iteration in Algorithm 2 has the same computational complexity if we iterate over every variable, but we can also stop the algorithm after obtaining $t$ variables. In this case, the total number of iterations $\gamma$ is $t = d[1 - (1 - \beta)^\gamma]$ which will require $\sum_{i=0}^{\gamma-1} d(1 - \beta)^i = d[1 - (1 - \beta)^\gamma]/\beta = t/\beta$ operations. Algorithm 2 thus takes $O(tn^3/\beta)$ time to discover $t$ variables

## 5 Experiments

### 5.1 Evaluation

We included two K-CDMs in Algorithm 1 by using Equation 3 or 4, which we will now denote as Proposed-F and Proposed-Z respectively. We compared Proposed-F and Proposed-Z with four feature ranking methods including BAHSIC, Relief-F and SVM-RFE. Rankings were normalized to compare variables with different sized Markov blankets as follows. If a continuous set of correct $MB(Y)$ variables were identified, then those variables were given the same rank. However, a break in the correct identification led to a higher rank. For example, if variables 2, 3 and 4 are in $MB(Y)$ while 1, 5, and 6 are not, then an output of 6,3,5,4,2,1 in ascending order would be converted to the ranking 5,4,3,2,2,1. The algorithm which provides a lower mean rank of $MB(Y)$ was then judged to perform better. In the example, the mean rank is 2.666, since the ranks of $MB(Y)$ are 4,2,2.

Next, we compared Algorithm 1 with three conditional dependence-based feature subset selection methods including IAMB, HITON-MB and MMMB by using the following accuracy measure:

$$A(X_c^\dagger, MB(Y)) = \frac{|X_c^\dagger \cap MB(Y)|}{|X_c^\dagger \cup MB(Y)|} * 100, \tag{5}$$

where $X^\dagger_c$ is the subset output from the conditional dependence algorithms or, for Proposed-F and Z, $X^\dagger_c$ is $X^\dagger$ clipped to the size of $MB(Y)$. For example, if variables 2, 3 and 4 are in $MB(Y)$ while 1, 5, and 6 are not, then an output of 6,3,5,4,2,1 from Algorithm 1 would be converted 4,2,1. Also, $|X^\dagger_c \cap MB(Y)|$ is the cardinality of the intersection of the subset $X^\dagger_c$ and the known $MB(Y)$ and $|X^\dagger_c \cup MB(Y)|$ is the cardinality of the union. Note that score *A* is equal to 100 when the algorithm outputs the exact $MB(Y)$. On the other hand, decreasing the cardinality of $X^\dagger_c$ by failing to identify parts of the $MB(Y)$ or increasing the cardinality of $X^\dagger_c$ by random guessing both decrease *A*.

### 5.2 Synthetic Datasets

Due to the debate presented by Lou and Obradovic (2010), we first evaluated the reliability of the dependence and conditional dependence measures in correctly identifying $MB(Y)$ under multiple conditions by comparing BAHSIC to Proposed-F and Proposed-Z (Figure 1). We compared these two algorithms because BAHSIC, Proposed-F and Proposed-Z have similar algorithmic structures but the former uses HSIC to measure dependence while the latter two use a K-CDM. We constructed synthetic Markov blankets containing 6 continuous variables (2 parents, 2 children, 2 spouses) by (1) generating the data points of 2 parents and 2 spouses by drawing from a Gaussian distribution with a standard deviation of 1, (2) summing the 2 parents and adding Gaussian noise with a standard deviation of 1 to create the data points of $Y$, (3) similarly summing the spouses and $Y$ and adding noise to create the data points of the 2 children. Thus, variables in $MB(Y)$ were connected by linear weights of 1. We then equipped BAHSIC, Proposed-F and Proposed-Z with linear kernels. In Figure 1, the solid lines represent the average ranking of $MB(Y)$ with the corresponding 95% confidence intervals shown as two dashed lines of the same color. For the first experiment, we introduced 10 extraneous variables drawn from a Gaussian distribution with a standard deviation of 1 to the original 7 variables (target plus 6 $MB(Y)$ variables) and varied the number of data points from 1 to 500. We found that BAHSIC performed better in the small sample size range (<75) but was then overtaken by Proposed-F and Proposed-Z. In order to understand this phenomenon, recall that the parents and children display an association to the target in this case whereas the spouses do not. As a result, BAHSIC cannot detect the 2 spouses and saturates at an average rank of 3, whereas Proposed-F and Proposed-Z continue to improve. For the second experiment, we raised the noise level throughout the entire dataset from 0 to 5 standard deviations while keeping the sample size constant at 70 corresponding to 10 data points for the target and each of the 6 variables in $\mathbf{MB}(Y)$. Proposed-F and Proposed-Z performed better up to about a noise standard deviation of 1, suggesting that it may be more reliable to search for $\mathbf{MB}(Y)$ using dependence measures instead of conditional dependence measures in high noise situations. This is expected, since the spouses need a common child to be predictive (Guyon et al., 2007), and thus their signal may be easily erased with noise. Next, we re-connected the 17 variables with 1 to 100 edges, again with a sample size of 70. We also varied the number of extraneous variables from 1 to 128 with the same sample size. Finally, we changed the value of the linear weights from 0.1 to 2. Proposed-F and Proposed-Z outperformed BAHSIC in these last three experimental conditions across all values. Moreover, Proposed-F and Proposed-Z gave identical to near identical results in all of the 5 experiments; the difference was greatest in the extraneous variables experiment, but it was only by 2-3 ranks with 64 and 128 extraneous variables. These results suggest that both K-CDMs can perform better than dependence based methods in correctly identifying $\mathbf{MB}(Y)$ when the noise level is low enough and the sample size is large enough.

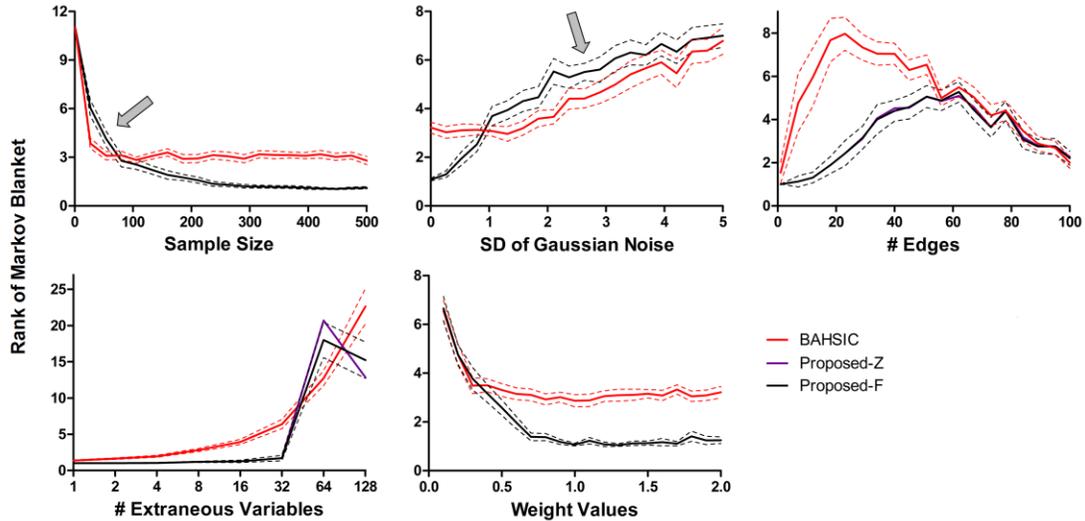

Figure 1: Results from synthetic datasets assessing the accuracy of dependency and conditional dependency-based methods in detecting *MB(Y)* by comparing Proposed-F and Proposed-Z to BAHSIC. Solid lines represent the average rank of the Markov blanket and dotted lines represent the 95% confidence interval. Proposed-F and Proposed-Z outperform BAHSIC except in low sample size and high noise conditions as indicated by the arrows. Moreover, BAHSIC consistently fails to identify the spouses by saturating at a rank of 3, whereas the proposed algorithm does not.

We compared Proposed-F and Proposed-Z to IAMB with Fisher's Z-test for the second set of synthetic experiments (Figure 2). We wanted to compare the accuracy of directly performing backward elimination on the dataset using a K-CDM instead of first performing statistical testing with a forward selection step. The HITON-MB and MMMB algorithms were not included, since they are data efficient modifications of IAMB which do not help in better assessing the impact of the forward selection step; however, these two algorithms are included in the next subsection. We found that Proposed-F and Proposed-Z outperformed IAMB across all 5 experiments, since the forward selection step may prevent IAMB from considering all multivariate combinations when discovering $MB(Y)$. Note that IAMB performs particularly poorly in the edges experiment as the Markov blanket size grows because statistical testing becomes unreliable with a fixed sample size. On the other hand, Proposed-F and Proposed-Z overcome this problem by not relying on statistical testing.

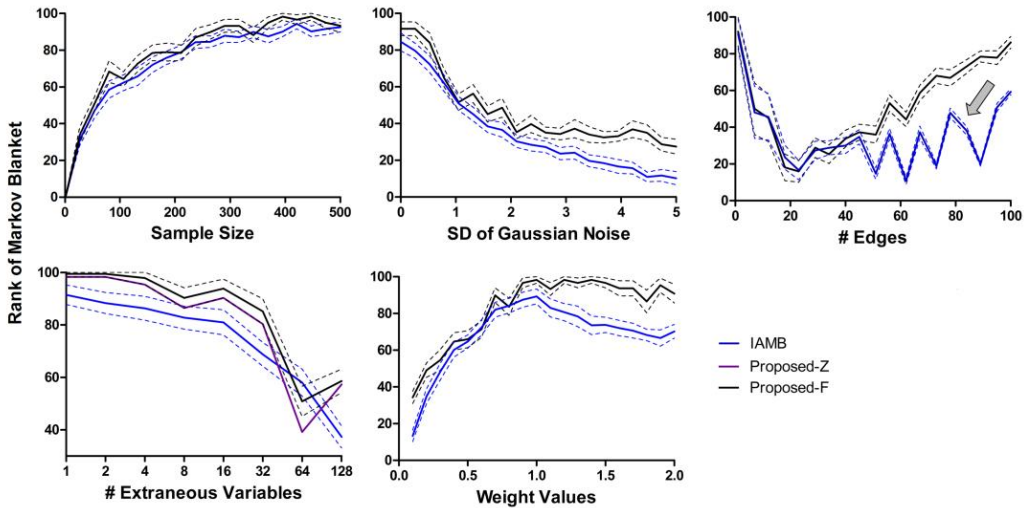

Figure 2: Results from synthetic datasets assessing the impact of a forward selection step by comparing Proposed-F and Proposed-Z to IAMB. Solid lines and dotted lines represent the average value of the accuracy measure in Equation 5 and 95% confidence intervals, respectively. Proposed-F and Proposed-Z outperform IAMB in all tested conditions. Notice that IAMB performs poorly in the edges experiment as indicated by the arrow, since statistical testing becomes unreliable with a growing $\mathbf{MB}(Y)$ size but fixed sample size.

### 5.3   Expert-Designed Models and Real-World Datasets

We used three publicly available expert-designed Bayesian network models including Alarm (36 variables), Child (20), and Insurance (27) as well as two real-world datasets including CYTO (11; Sachs et al., 2005) and the U.S. Linked Infant Birth and Death Dataset from 1991 (87; Mani & Cooper, 1999). CYTO is a dataset of T-lymphocyte protein-protein interactions, and Infant is a dataset of clinical outcomes and decisions regarding infant births; in both of these, portions of $MB(Y)$ have been experimentally verified and confirmed by experts. We appropriately incorporated RBF kernels with sigma set to the median distance between data points in all kernel methods to detect discrete non-linear patterns. The IAMB, HITON-MB, and MMMB algorithms were implemented with the $G^2$ test for discrete data. We iterated over all variables to obtain the mean rank and accuracy scores over different sample sizes. Results are shown in Figures 3 and 4 for the expert-designed models and real-world datasets, respectively.

The results show that both Proposed-F and Proposed-Z outperform other feature ranking and subset selection methods in correctly identifying $MB(Y)$ with larger sample sizes in the datasets of expert-designed models. Notice that the dependency based method BAHSIC plateaus at a relatively small sample size, but the proposed algorithm's performance continues to improve with larger sample sizes. These results held when using either the method from Fukumizu et al. (2009) or Zhang et al. (2011) as the K-CDM. For the real-world datasets, Proposed-F and Proposed-Z outperformed all other conditional dependence-based algorithms. The results are less clear when comparing against the ranking algorithms in CYTO, since no algorithm consistently outperformed the others, but we observed that the proposed algorithm significantly outperformed

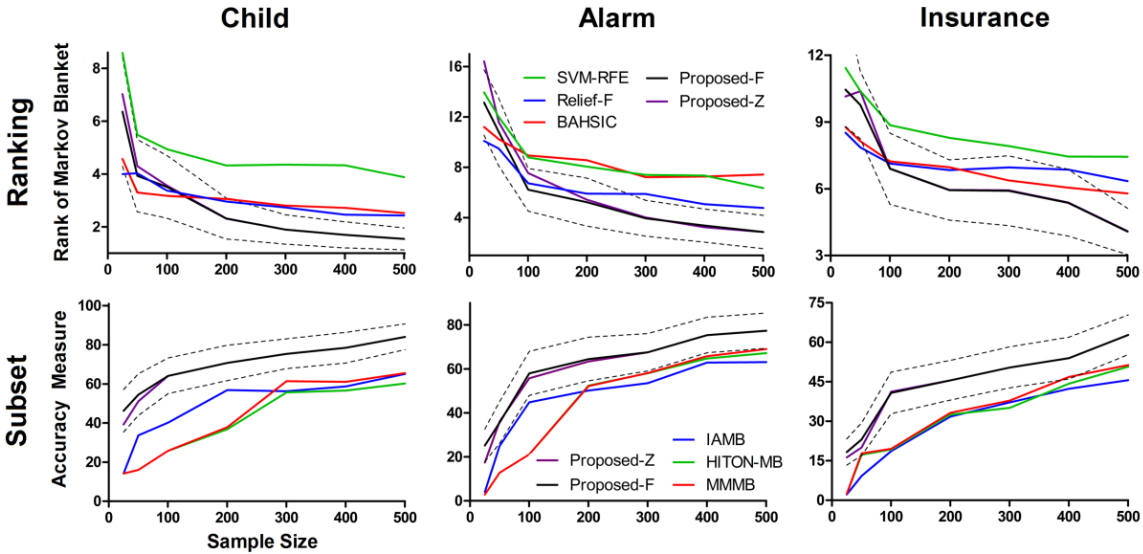

Figure 3: Results from datasets created from expert-designed models. Solid lines and dotted lines again represent the average ranks of the Markov blanket or the average value of the accuracy measure in Equation 5 and 95% confidence intervals, respectively. Proposed-F and Proposed-Z outperform all ranking methods across the larger sample sizes and subset selection methods across all of the sample sizes.

Relief-F on occasion. For Infant, the proposed algorithm was outperformed by BAHSIC, since the Markov blankets in this dataset only contain parents and children; in this situation, kernel-based dependency methods may perform better, as we observed in the synthetic experiments.

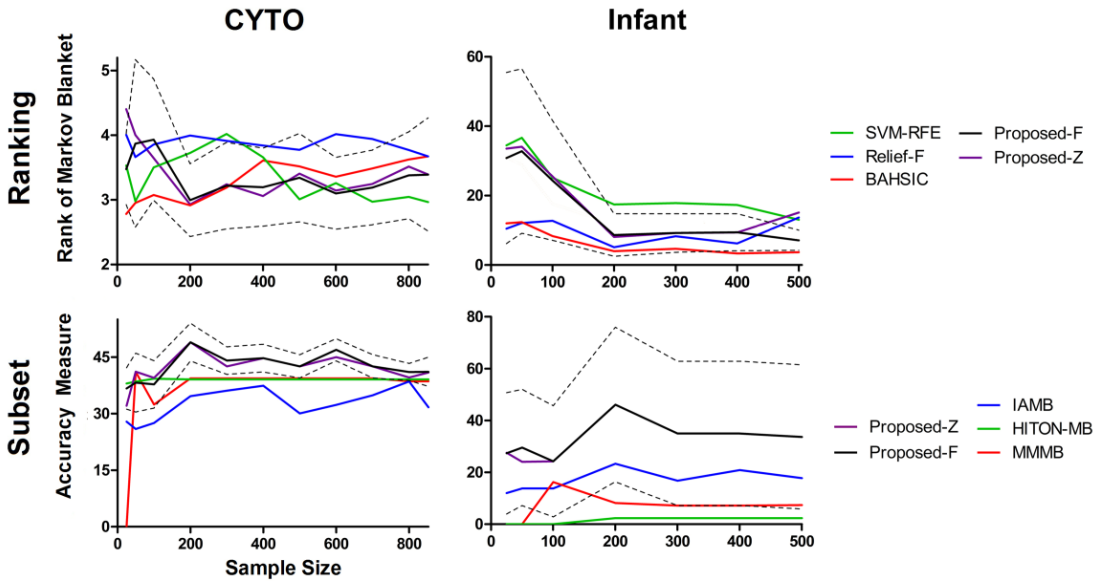

Figure 4: Results from real-world datasets. Proposed-F and Proposed-Z outperform all subset selection methods across all of the sample sizes. However, both methods are consistently outperformed by BAHSIC in Infant.

## 6 Conclusion

We introduced a feature ranking algorithm that is useful for discovering $MB(Y)$. The algorithm uses a K-CDM to eliminate variables using backward elimination. Overall, the method exhibits superior performance in synthetic data and in real datasets on average when compared to several feature ranking and subset selection methods.

**Acknowledgements**

We thank Dr. Subramani Mani for providing the U.S. Linked Infant Birth and Death 1991 dataset. This research was funded by the National Library of Medicine grant T15 LM007059-24 to the University of Pittsburgh Biomedical Informatics Training Program and the National Institute of General Medical Sciences grant T32 GM008208 to the University of Pittsburgh Medical Scientist Training Program.

**References**

C. F. Aliferis, I. Tsamardinos, and A. Statnikov. HITON: a novel Markov blanket algorithm for optimal variable selection. *AMIA 2003 Annual Symposium Proceedings*, 21–25, 2003.

C. Cortes, M. Mohri, and A. Rostamizadeh. Algorithms for learning kernels based on centered alignment. *The Journal of Machine Learning Research*, 13:795-828, 2012.

N. Friedman, and D. Koller. *Probabilistic Graphical Models: Principles and Techniques*. The MIT Press, 2009.

K. Fukumizu, F. R. Bach, and M. I. Jordan. Kernel dimension reduction in regression. *Annals of Statistics*, 37(5):1871–1905, 2009.

A. Gretton, R. Herbrich, A. Smola, O. Bousquet, and B. Schölkopf. Kernel methods for measuring independence. *Journal of Machine Learning Research*, 6:2075–2129, 2005.

I. Guyon, C. F. Aliferis, and A. Elisseeff. *Computational Methods of Feature Selection*, chapter Causal Feature Selection. Chapman and Hall, 2007.

P. Hoyer, D. Janzing, J. Mooij, J. Peters, and B. Schölkopf. Nonlinear causal discovery with additive noise models. *Neural Information Processing Systems*, 2008.

Q. Lou and Z. Obradovic. Feature selection by approximating the Markov blanket in a kernel induced space. *European Conference on Artificial Intelligence*, 797-802, 2010.

S. Mani and G. F. Cooper. A study in causal discovery from population-based infant birth and death records. *AMIA Annual Fall Symposium*, 319, 1999.

J. M. Peña, J. Björkegren, and J. Tegner. Scalable, efficient and correct learning of Markov boundaries under the faithfulness assumption. *Eighth European Conference on Symbolic and Quantitative Approaches to Reasoning under Uncertainty*, 136–147, 2005.

K. Sachs, O. Perez, D. Pe'er, D. Lauenburger, and G. Nolan. Causal protein-signaling networks derived from multiparameter single-cell data. *Science*, 308, 2005.

S. Shimizu, P.O. Hoyer, A. Hyvärinen, and A.J. Kerminen. A linear, non-Gaussian acyclic model for causal discovery. *Journal of Machine Learning Research*, 7:2003-2030, 2006.

L. Song, J. Bedo, K. M. Borgwardt, A. Gretton, and A. Smola. Gene selection via the BAHSIC family of algorithms. *Bioinformatics*, 23(13):490–498, 2007.

P. Spirtes. An anytime algorithm for casual inference. *Proceedings of the Eighth International Workshop on Artificial Intelligence and Statistics*, 213–221, 2001.

P. Spirtes, C. Glymour, and R. Scheines. *Causation, Prediction, and Search*. The MIT Press, 2$^{nd}$ edition, 2000.

A. Statnikov, N. Lytkin, J. Lemeire and C. F. Aliferis. Algorithms for discovery of multiple Markov boundaries. *Journal of Machine Learning Research* 14, 499-566, 2013.

I. Tsamardinos and C. F. Aliferis. Towards principled feature selection: relevancy, filters and wrappers. *Proceedings of the Ninth International Workshop on Artificial Intelligence and Statistics*, 2003.

I. Tsamardinos, L. E. Brown, and C. F. Aliferis. The max-min hill-climbing Bayesian network structure learning algorithm. Machine Learning, 65(1):31–78, 2006.

K. Zhang, J. Peters, D. Janzing, and B. Schölkopf. Kernel-based conditional independence test and application to causal discovery. *Proceedings of Uncertainty in Artificial Intelligence*, 804-813, 2011.